\PassOptionsToPackage{table}{xcolor}
\documentclass[sigconf,nonacm]{acmart}

\usepackage{balance}
\usepackage{booktabs}
\usepackage{array}
\usepackage{multirow}
\usepackage{adjustbox}  
\usepackage{tcolorbox}
\usepackage{tikz}
\usetikzlibrary{positioning,arrows.meta,calc,fit,backgrounds,decorations.pathreplacing}
\makeatletter

\definecolor{rq1}{HTML}{6B4C9A}
\definecolor{rq2}{HTML}{A5620C}

\newtcbox{\rq@badge}[1]{on line, nobeforeafter, colback=rq#1, colframe=rq#1,
  coltext=white, boxrule=0pt, arc=1.2pt, boxsep=0pt,
  left=2.2pt, right=2.2pt, top=1pt, bottom=0.8pt,
  fontupper=\sffamily\bfseries\footnotesize}
\DeclareRobustCommand{\RQ}[1]{\rq@badge{#1}{RQ#1}}

\newsavebox{\rq@savebox}
\newtcolorbox{rqbox}[1]{colback=rq#1!7!white, colframe=rq#1, boxrule=0.6pt, arc=2pt,
  boxsep=0pt, left=4pt, right=5pt, top=3.5pt, bottom=3.5pt,
  before skip=5pt plus 2pt, after skip=5pt plus 2pt,
  before upper={\sbox\rq@savebox{\RQ{#1}\hspace{0.5em}}%
    \hangindent=\wd\rq@savebox \hangafter=1 \noindent\usebox\rq@savebox}}

\newcommand{\rq@note}[1]{\unskip\nobreak\hfil\penalty50\hskip1.5em\hbox{}\nobreak\hfill
  {\color{black!60}#1}\parfillskip=0pt\finalhyphendemerits=0\par}

\newcommand{\defineRQ}[2]{\@namedef{rq@text@#1}{#2}}
\newcommand{\showRQ}[2][]{\begin{rqbox}{#2}\@nameuse{rq@text@#2}%
  \if\relax\detokenize{#1}\relax\else\rq@note{#1}\fi\end{rqbox}}

\makeatother

\defineRQ{1}{Do decision models rerank better than the untuned model or a reranker fine-tune of the same backbone?}
\defineRQ{2}{Are their relevance probabilities better calibrated than those of the untuned model and the reranker
  fine-tune, and is that calibration of any use?}

\usepackage{placeins}  
\newsavebox{\widetablebox}
\newcommand{\inputwidetable}[1]{%
  \begingroup
  \let\origtabular\tabular \let\endorigtabular\endtabular
  \renewenvironment{tabular}[1]{\begin{lrbox}{\widetablebox}\origtabular{##1}}%
    {\endorigtabular\end{lrbox}\resizebox{\linewidth}{!}{\usebox{\widetablebox}}}%
  \input{#1}%
  \endgroup}

\colorlet{rowbase}{black!14}
\colorlet{rowdecision}{red!20}
\colorlet{rowreranker}{blue!18}

\newcommand{\laya}{Laya}
\newcommand{\jev}{Jev}

\setcopyright{none}

\begin{document}

\title{System One Models Are Not Better Rerankers}

\author{Michael Dinzinger}
\affiliation{%
  \institution{University of Passau}
  \city{Passau}
  \country{Germany}}
\email{michael.dinzinger@uni-passau.de}

\author{Kanishka Ghosh Dastidar}
\affiliation{%
  \institution{Interdisciplinary Transformation University Austria (IT:U)}
  \city{Linz}
  \country{Austria}}
\email{kanishka.ghosh-dastidar@it-u.at}

\author{Michael Granitzer}
\affiliation{%
  \institution{University of Passau}
  \city{Passau}
  \country{Germany}}
\affiliation{%
  \institution{Interdisciplinary Transformation University Austria (IT:U)}
  \city{Linz}
  \country{Austria}}
\email{michael.granitzer@uni-passau.de}

\renewcommand{\shortauthors}{Dinzinger et al.}

\begin{abstract}
System One models, or decision models, answer a question about an input with a calibrated probability instead of
generated text. Within weeks of the first release, in September 2026, they were used as rerankers. We test whether
the decision fine-tuning behind them makes a better reranker than plain reranker fine-tuning of the same model. For
six Qwen backbones from 0.6B to 27B parameters and ten TREC DL and BEIR collections, we compare the untuned model,
asked zero-shot, with its decision fine-tunes and its reranker fine-tunes, two of which we train ourselves.
In our work, we draw the conclusion that decision
fine-tuning never makes the better reranker. From 0.6B to 9B, reranker fine-tuning wins on every backbone by 5 to 29
points nDCG@10, also on zero-shot collections, and at 27B the two tie. Up to 9B, decision
fine-tuning adds nothing over simply asking the untuned model.
Moreover, we find that the calibration advantage of decision models depends
on the question. Asked whether a passage answers the query, they are better calibrated than the published reranker
fine-tunes from 4B on. Asked a slightly different question from prior work, they lose this advantage.
Lastly, calibration does not make up for a worse ranking: up to 9B, a reranker fine-tune keeps the
better passages at a fixed threshold. If you need a reranker, fine-tune a reranker.

\end{abstract}


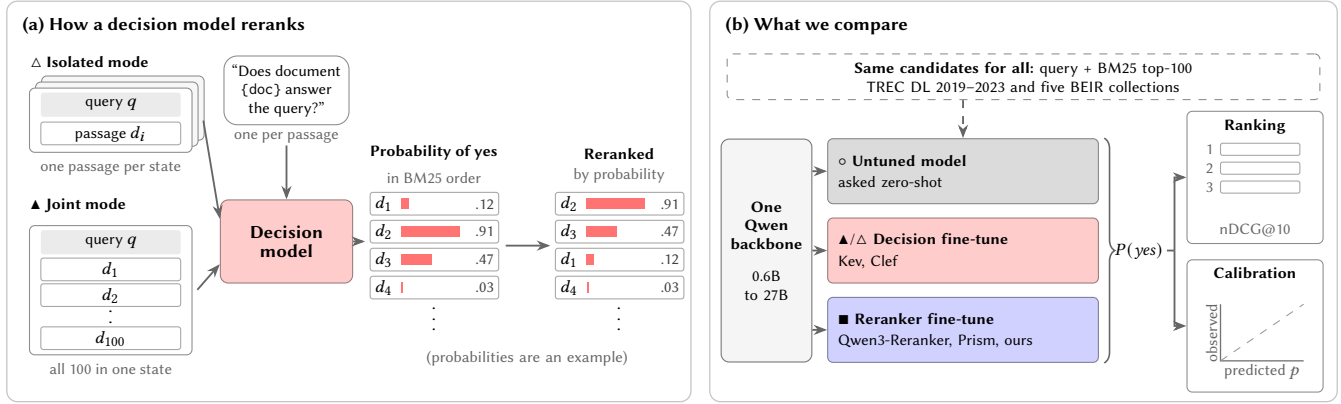
\begin{teaserfigure}
\centering
\begin{tikzpicture}[
    font=\sffamily\scriptsize,
    >={Stealth[length=1.8mm]},
    every node/.style={inner sep=2pt},
    flow/.style={->, line width=0.7pt, black!60},
    panel/.style={draw=black!22, rounded corners=4pt, line width=0.5pt},
    ptitle/.style={font=\sffamily\footnotesize\bfseries, anchor=north west, inner sep=0pt},
    head/.style={font=\sffamily\scriptsize\bfseries, inner sep=0pt},
    note/.style={black!60, align=center, inner sep=1pt},
    state/.style={draw=black!45, fill=white, rounded corners=2pt},
    qline/.style={fill=black!8, rounded corners=1pt, minimum width=1.85cm, minimum height=0.32cm, inner sep=0pt},
    card/.style={draw=black!35, fill=white, rounded corners=1pt, minimum width=1.85cm, minimum height=0.3cm,
      inner sep=0pt},
    variant/.style={draw=#1!60!black, fill=#1, rounded corners=2pt, minimum width=3.5cm, minimum height=0.86cm,
      text width=3.3cm, align=left, inner xsep=4pt},
    evalbox/.style={draw=black!45, fill=white, rounded corners=2pt, minimum width=1.8cm, minimum height=1.75cm},
  ]

  \draw[panel] (0,0) rectangle (9.05,5.3);
  \node[ptitle] at (0.2,5.1) {(a) How a decision model reranks};

  \node[head, anchor=west] at (0.3,4.5) {$\triangle$ Isolated mode};
  \foreach \k in {2,1}{
    \draw[state, fill=black!3] ($(0.3,3.3)+(\k*0.08,\k*0.08)$) rectangle ++(2.15,0.85);
  }
  \draw[state] (0.3,3.3) rectangle (2.45,4.15) coordinate[pos=0.5] (iso);
  \node[qline] at (1.375,3.93) {query $q$};
  \node[card] at (1.375,3.55) {passage $d_i$};
  \node[note, anchor=north] at (1.375,3.22) {one passage per state};

  \node[head, anchor=west] at (0.3,2.6) {$\blacktriangle$ Joint mode};
  \draw[state] (0.3,0.62) rectangle (2.45,2.33) coordinate[pos=0.5] (joint);
  \node[qline] at (1.375,2.11) {query $q$};
  \node[card] at (1.375,1.75) {$d_1$};
  \node[card] at (1.375,1.41) {$d_2$};
  \node[inner sep=0pt] at (1.375,1.16) {$\vdots$};
  \node[card] at (1.375,0.84) {$d_{100}$};
  \node[note, anchor=north] at (1.375,0.55) {all 100 in one state};

  \node[draw=black!45, fill=white, rounded corners=5pt, align=center, inner sep=3pt]
    (question) at (3.7,4.12) {``Does document\\\texttt{\{doc\}} answer\\the query?''};
  \node[note, anchor=north] (each) at ($(question.south)+(0,-0.04)$) {one per passage};
  \node[draw=rowdecision!60!black, fill=rowdecision, rounded corners=3pt, minimum width=1.75cm,
    minimum height=1.1cm, align=center] (model) at (3.7,2.12) {\footnotesize\textbf{Decision}\\
    \footnotesize\textbf{model}};
  \draw[flow] (each.south) -- (model.north);
  \draw[flow] (2.6,3.72) -- ($(model.west)+(0,0.3)$);
  \draw[flow] (2.52,1.48) -- ($(model.west)+(0,-0.3)$);

  \def\barmax{0.85}
  \def\probcard#1#2#3#4{%
    \node[card, minimum width=1.7cm, anchor=west] at (#1,#2) {};
    \node[anchor=west, inner sep=0pt] at ($(#1,#2)+(0.07,0)$) {#3};
    \fill[red!55] ($(#1,#2)+(0.42,-0.07)$) rectangle ++(#4*\barmax,0.14);
    \node[anchor=east, inner sep=0pt, black!75] at ($(#1,#2)+(1.65,0)$) {.\pgfmathparse{int(round(#4*100))}%
      \ifnum\pgfmathresult<10 0\fi\pgfmathresult};
  }
  \node[head, anchor=south] at (5.65,3.2) {Probability of yes};
  \node[note, anchor=south] at (5.65,2.85) {in BM25 order};
  \probcard{4.8}{2.62}{$d_1$}{0.12}
  \probcard{4.8}{2.25}{$d_2$}{0.91}
  \probcard{4.8}{1.88}{$d_3$}{0.47}
  \probcard{4.8}{1.51}{$d_4$}{0.03}
  \node[inner sep=0pt] at (5.65,1.2) {$\vdots$};
  \draw[flow] (model.east) -- (4.74,2.12);

  \node[head, anchor=south] at (8.1,3.2) {Reranked};
  \node[note, anchor=south] at (8.1,2.85) {by probability};
  \probcard{7.25}{2.62}{$d_2$}{0.91}
  \probcard{7.25}{2.25}{$d_3$}{0.47}
  \probcard{7.25}{1.88}{$d_1$}{0.12}
  \probcard{7.25}{1.51}{$d_4$}{0.03}
  \node[inner sep=0pt] at (8.1,1.2) {$\vdots$};
  \draw[flow] (6.6,2.065) -- (7.19,2.065);
  \node[note, anchor=north, text width=4.2cm] at (6.88,0.72) {(probabilities are an example)};

  \begin{scope}[xshift=9.3cm]
  \draw[panel] (0,0) rectangle (8.3,5.3);
  \node[ptitle] at (0.2,5.1) {(b) What we compare};

  \node[draw=black!45, dashed, rounded corners=2pt, align=center, text width=7.7cm, inner ysep=3pt]
    (input) at (4.15,4.3) {\textbf{Same candidates for all:} query + BM25 top-100\\TREC DL 2019--2023 and five BEIR collections};

  \node[draw=black!55, fill=black!4, rounded corners=3pt, minimum height=2.95cm, minimum width=1.2cm,
    text width=1.1cm, inner xsep=1pt, align=center] (backbone) at (0.75,2.0) {\textbf{One}\\\textbf{Qwen}\\\textbf{backbone}\\[5pt]
    0.6B\\to 27B};

  \node[variant=rowbase] (base) at (3.35,3.05)
    {$\circ$~\textbf{Untuned model}\\[1pt]asked zero-shot};
  \node[variant=rowdecision] (decision) at (3.35,2.0)
    {$\blacktriangle$/$\triangle$~\textbf{Decision fine-tune}\\[1pt]Kev, Clef};
  \node[variant=rowreranker] (reranker) at (3.35,0.95)
    {$\blacksquare$~\textbf{Reranker fine-tune}\\[1pt]Qwen3-Reranker, Prism, ours};
  \foreach \v in {base, decision, reranker}{
    \draw[flow] (backbone.east |- \v.west) -- (\v.west);
  }
  \draw[flow, dashed] (input.south -| base.north) -- (base.north);

  \draw[decorate, decoration={brace, amplitude=4pt}, line width=0.6pt, black!60]
    ($(base.north east)+(0.07,0)$) -- ($(reranker.south east)+(0.07,0)$) coordinate[midway] (brace);
  \node[anchor=west, inner sep=1pt, align=center] (pyes) at ($(brace)+(0.1,0)$) {$P(\mathit{yes})$};

  \node[evalbox, anchor=north east] (rank) at (8.12,3.85) {};
  \node[head, anchor=north] at ($(rank.north)+(0,-0.1)$) {Ranking};
  \foreach \i in {1,2,3}{
    \node[inner sep=0pt, black!60, anchor=east] at ($(rank.north west)+(0.36,-0.26-\i*0.25)$) {\i};
    \draw[draw=black!35, fill=white, rounded corners=0.8pt]
      ($(rank.north west)+(0.45,-0.34-\i*0.25)$) rectangle ++(1.05,0.16);
  }
  \node[note, anchor=south] at ($(rank.south)+(0,0.06)$) {nDCG@10};

  \node[evalbox, anchor=south east] (cal) at (8.12,0.12) {};
  \node[head, anchor=north] at ($(cal.north)+(0,-0.1)$) {Calibration};
  \begin{scope}[shift={($(cal.south west)+(0.45,0.42)$)}]
    \draw[black!55, line width=0.5pt] (0,0.75) -- (0,0) -- (1.1,0);
    \draw[black!40, dashed] (0,0) -- (1.1,0.75);
    \node[note, anchor=north] at (0.55,0) {predicted $p$};
    \node[note, anchor=south, rotate=90] at (0,0.375) {observed};
  \end{scope}

  \draw[line width=0.7pt, black!60] (pyes.east) -- ++(0.1,0) coordinate (fork);
  \draw[flow] (fork) |- (rank.west);
  \draw[flow] (fork) |- (cal.west);
  \end{scope}
\end{tikzpicture}

  \caption{How System One models rerank and what we compare. (a)~For each BM25
  candidate, a decision model is asked whether the passage answers the query. The candidates are sorted by the probability of ``yes''. The
  input of the model (state) holds one passage (isolated mode) or all 100 candidates (joint mode). (b)~On one Qwen backbone, the untuned
  model, its decision fine-tunes and its reranker fine-tunes rerank the same candidates and are compared on ranking
  and on calibration.}
  \Description{Two panels. Panel a: a query with one passage per state (isolated mode) or with all 100 candidate
  passages in one state (joint mode) goes into a decision model together with the question ``Does document doc answer
  the query?'', asked once per passage. The model returns a probability of yes for every passage, shown as bars on
  example passages in BM25 order, and sorting by this probability gives the reranked list. Panel b: one Qwen backbone
  of 0.6B to 27B parameters branches into three variants: the untuned model asked zero-shot, a decision fine-tune (Kev,
  Clef) and a reranker fine-tune (Qwen3-Reranker, Prism, ours). All three read the same BM25 top-100 of TREC DL
  2019--2023 and five BEIR collections and return a probability of yes, which is evaluated for ranking (nDCG@10) and
  for calibration (predicted probability against observed relevance).}
  \label{fig:teaser}
  \vspace{0.3cm}
\end{teaserfigure}

\maketitle

\section{Introduction}
\label{sec:introduction}

Cross-encoders are a well-established way to rerank the candidates of a first-stage
retriever~\cite{Nogueira2019,Nogueira2020monoT5}, and open large language models fine-tuned for ranking are the
current state of the art~\cite{Ma2023RankLLaMA,Zhang2025Qwen3Embedding,Zhang2026Prism}. \emph{System One models},
which we call decision models, promise something else. They answer a question about an input with a calibrated
probability instead of generated text. The name refers to fast, intuitive thinking, Kahneman's
``System~1''~\cite{Kahneman2011,TypeSafe2026Jev}. TypeSafe AI released the first of them, \jev{}, on 15 September
2026, trained with an unpublished method called RLCD (reinforcement learning for calibrated
decisions)~\cite{TypeSafe2026Jev}. Within weeks, \jev{} was reported to be a competitive
reranker~\cite{S1Rank,JevRerank} and was evaluated or integrated in retrieval
systems~\cite{Elastic2026,LanceDB2026,Hindsight2026}. Open decision models followed: Kev~\cite{Kev} and Cloudflare's
Clef~\cite{Clef}. They are fine-tuned from Qwen backbones, the same backbones that dedicated rerankers such as
Qwen3-Reranker~\cite{Zhang2025Qwen3Embedding} and Prism~\cite{Zhang2026Prism} are built on.

This makes a clean test possible, and the reports so far have not run it. They compare \jev{} with rerankers built
on other, mostly smaller models, so they cannot tell how much of its effectiveness comes from decision fine-tuning
and how much from the underlying language model. We therefore compare decision fine-tuning with plain reranker
fine-tuning of the same backbone. Calibration needs a test of its own. A ranking uses only the order of the scores,
so a model can rank well with badly calibrated probabilities, and the other way round. Calibration matters when a
score is read on an absolute scale: to decide how many passages to pass on to a language model, to drop passages
that do not answer the query, or to estimate how many relevant passages there are. We therefore also test whether the
advertised calibration holds for relevance and whether it helps in such use cases:

\showRQ{1}
\showRQ{2}

Figure~\ref{fig:teaser} shows how a decision model reranks and what we compare. For six Qwen backbones from 0.6B to
27B parameters, we evaluate the untuned model, asked zero-shot, its decision fine-tunes (Kev and Clef) and its
reranker fine-tunes, including two we train ourselves on MS~MARCO alone (at 9B as a control, at 27B where none has
been released). All of them rerank the same BM25 candidates of TREC DL 2019--2023 and five BEIR collections, and all
return a probability of relevance per passage. Decision and untuned models are asked whether the passage answers the
query. For comparison, we repeat all their runs with the looser question of S1Rank~\cite{S1Rank}, an earlier report
on \jev{} as a reranker. We find:
\begin{enumerate}
  \item \emph{Reranker fine-tuning beats decision fine-tuning.} From 0.6B to 9B, the reranker fine-tune ranks better
  on every backbone, by 5 to 29 points nDCG@10. The difference is significant on 48 of the 60 pairs of reranker
  fine-tune and collection, never significantly in favour of the decision fine-tune, and it holds on collections that
  no fine-tune was trained on. At 27B, the two tie. On no backbone does decision fine-tuning make the better reranker.
  \item \emph{Up to 9B, decision fine-tuning adds nothing over the untuned model.} It stays within $-2.5$ and $+0.7$
  points of simply asking the untuned model the same question, while reranker fine-tuning adds 6 to 30 points.
  \item \emph{The calibration advantage depends on the question.} Asked whether a passage answers the query, decision
  models from 4B on are better calibrated than the published reranker fine-tunes. Asked S1Rank's question, they lose
  this advantage while their ranking hardly moves, and the untuned models gain as much from the wording. Calibration
  also does not make up for a worse ranking: up to 9B, a reranker fine-tune keeps the better passages at a fixed
  threshold.
\end{enumerate}
Our position is straightforward and runs counter to the recent trend: System One models are not better rerankers.%
\footnote{Code, model outputs (a probability per query and passage) and all results are public: \url{https://github.com/KanishkaGD/decision-rerank}}

\section{Related Work}
\label{sec:related-work}

\paragraph{LLM-based rerankers.}
Neural rerankers are trained to score query--passage pairs from relevance labels, first with encoder or
sequence-to-sequence models~\cite{Nogueira2019,Nogueira2020monoT5} and more recently with open large language
models~\cite{Ma2023RankLLaMA,Zhang2025Qwen3Embedding,Zhang2026Prism}. Qwen3-Reranker and Prism read relevance from the
probability of ``yes'' against ``no'' after a prompt that asks whether the document meets the query, which makes their
output directly comparable to that of a decision model. Other approaches rerank the candidates jointly, either by
generating a ranked list~\cite{Sun2023RankGPT,Pradeep2023RankZephyr} or by letting passages attend to each other, as in
the Set-Encoder~\cite{Schlatt2025SetEncoder}. Large language models can also rank without any reranking-specific
fine-tuning, through relevance prompting~\cite{Qin2024PRP,Zhuang2024BeyondYesNo}, but their effectiveness is sensitive
to how the prompt is worded~\cite{Sun2025PromptVariations}. Abdallah et~al.~\cite{Abdallah2025HowGood} compare many of
these rerankers on the candidates we use.

\paragraph{Decision models in retrieval.}
Besides \jev{}~\cite{TypeSafe2026Jev}, Kev~\cite{Kev,KevCards}, Clef~\cite{Clef} and \laya{}~\cite{Laya} are open
decision models. The evidence on decision models for retrieval is so far limited to reports that are not
peer-reviewed: \jev{} reranking BM25 candidates~\cite{S1Rank,JevRerank,LanceDB2026}, as a reranker in production
search~\cite{Elastic2026,Hindsight2026}, and the open \laya{} as a relevance gate before recall~\cite{Laya,Smriti2026}. These reports
compare \jev{} with existing rerankers, which are built on other models. None compares decision fine-tuning with
reranker fine-tuning of the same backbone, and none asks whether the calibration of the probabilities pays off.

\paragraph{Calibration of relevance scores.}
Neural networks tend to be overconfident, and temperature scaling after training is a simple
remedy~\cite{Guo2017Calibration}. Neural rankers are no exception~\cite{Penha2021Calibration}. Common ranking losses
are invariant to shifting the scores of a query, so rankers trained with them can order passages well without being
calibrated~\cite{Penha2021Calibration,Yan2022ScaleCalibration}; scores can be calibrated during or after
training~\cite{Guo2017Calibration,Yu2025ExplainThenRank}. Calibrated probabilities matter wherever a score is read on
an absolute scale, for example to cut off a ranked list~\cite{Bahri2020Choppy}. This is the use we test in RQ2.

\section{Decision Models as Rerankers}
\label{sec:decision-models}

\begin{table*}[t]
\caption{Untuned model, decision fine-tunes and reranker fine-tunes of the same backbone (nDCG@10,
$\times 100$, BM25 top-100). Untuned and decision models are asked whether the passage answers the query.
Avg: over all ten collections; ZS: mean over the collections without a training split in any
fine-tune's data (TREC-COVID, Touch\'e, DBPedia). Per backbone and column: best in \texttt{bold}, second best
\underline{underlined}. Training: $\circ$~untuned model, $\blacktriangle$/$\triangle$~decision fine-tune (joint/isolated), $\blacksquare$~reranker fine-tune (rows shaded grey, red, blue);
\textit{Ours}: our reranker fine-tunes.
Paired $t$-test, Holm-corrected over
the ten collections, $p < 0.05$: $^\ast$ reranker fine-tune vs.\ the decision fine-tune with the best Avg,
$^\dagger$ that decision fine-tune vs.\ the untuned model, $^\ddagger$ joint vs.\ isolated.}
\label{tab:triplets}
\centering
\scriptsize
\setlength{\tabcolsep}{3pt}
\begin{adjustbox}{max width=\textwidth}
\begin{tabular}{cllrrrrrrrrrrrr}
\toprule
 & & & \multicolumn{5}{c}{TREC DL} & \multicolumn{5}{c}{BEIR} & \multicolumn{2}{c}{Mean} \\
\cmidrule(lr){4-8}\cmidrule(lr){9-13}\cmidrule(lr){14-15}
 & System & Mode & DL19 & DL20 & DL21 & DL22 & DL23 & COVID & NFC & SciF & Touch\'e & DBP & Avg & ZS \\
\midrule
\multicolumn{15}{l}{\textit{Qwen3-0.6B}} \\
\rowcolor{rowbase}$\circ$ & Qwen3-0.6B &  & \underline{43.3} & \underline{32.7} & 37.7 & 22.1 & 17.9 & 50.9 & 15.1 & 14.2 & \underline{16.8} & 15.3 & 26.6 & 27.7 \\
\rowcolor{rowdecision}$\blacktriangle$ & Kev-0.6B & joint & 23.3\rlap{$^{\ddagger}$} & 15.1\rlap{$^{\ddagger}$} & 20.4\rlap{$^{\ddagger}$} & 12.8\rlap{$^{\ddagger}$} & 12.1\rlap{$^{\ddagger}$} & 36.8\rlap{$^{\ddagger}$} & 12.4\rlap{$^{\ddagger}$} & 4.5\rlap{$^{\ddagger}$} & 5.2\rlap{$^{\ddagger}$} & 8.4\rlap{$^{\ddagger}$} & 15.1 & 16.8 \\
\rowcolor{rowdecision}$\triangle$ & Kev-0.6B & isolated & 35.3 & 30.5 & \underline{40.1} & \underline{22.8} & \underline{19.0} & \underline{55.9} & \underline{16.5} & \underline{15.5} & 14.1 & \underline{22.9}\rlap{$^{\dagger}$} & \underline{27.3} & \underline{31.0} \\
\rowcolor{rowreranker}$\blacksquare$ & Qwen3-Reranker-0.6B &  & \textbf{69.5}\rlap{$^{\ast}$} & \textbf{64.0}\rlap{$^{\ast}$} & \textbf{65.2}\rlap{$^{\ast}$} & \textbf{46.4}\rlap{$^{\ast}$} & \textbf{38.8}\rlap{$^{\ast}$} & \textbf{86.2}\rlap{$^{\ast}$} & \textbf{36.4}\rlap{$^{\ast}$} & \textbf{76.6}\rlap{$^{\ast}$} & \textbf{35.7}\rlap{$^{\ast}$} & \textbf{42.7}\rlap{$^{\ast}$} & \textbf{56.1} & \textbf{54.9} \\
\midrule
\multicolumn{15}{l}{\textit{Qwen3-8B}} \\
\rowcolor{rowbase}$\circ$ & Qwen3-8B &  & 67.3 & \underline{63.7} & \underline{62.0} & 38.3 & \underline{39.0} & 77.6 & 32.7 & \underline{73.3} & \underline{25.5} & 31.6 & 51.1 & 44.9 \\
\rowcolor{rowdecision}$\blacktriangle$ & Kev-8B & joint & 62.7 & 55.2 & 59.1 & 35.5 & 34.3 & 53.1\rlap{$^{\ddagger}$} & 20.4\rlap{$^{\ddagger}$} & 17.9\rlap{$^{\ddagger}$} & 12.3\rlap{$^{\ddagger}$} & 36.2\rlap{$^{\ddagger}$} & 38.7 & 33.9 \\
\rowcolor{rowdecision}$\triangle$ & Kev-8B & isolated & \underline{69.0} & 63.1 & 61.9 & \underline{38.6} & 36.4 & \underline{80.6} & \underline{33.8} & 69.8 & 25.2 & \underline{39.0}\rlap{$^{\dagger}$} & \underline{51.7} & \underline{48.3} \\
\rowcolor{rowreranker}$\blacksquare$ & Qwen3-Reranker-8B &  & \textbf{73.2} & \textbf{66.8} & \textbf{67.0}\rlap{$^{\ast}$} & \textbf{51.2}\rlap{$^{\ast}$} & \textbf{45.2}\rlap{$^{\ast}$} & \textbf{84.0} & \textbf{37.3}\rlap{$^{\ast}$} & \textbf{79.9}\rlap{$^{\ast}$} & \textbf{43.1}\rlap{$^{\ast}$} & \textbf{46.2}\rlap{$^{\ast}$} & \textbf{59.4} & \textbf{57.8} \\
\midrule
\multicolumn{15}{l}{\textit{Qwen3.5-0.8B}} \\
\rowcolor{rowbase}$\circ$ & Qwen3.5-0.8B &  & \underline{46.0} & \underline{41.2} & \underline{52.1} & \underline{28.1} & \underline{27.9} & 59.2 & 18.3 & 27.0 & 16.4 & 22.3 & 33.9 & 32.6 \\
\rowcolor{rowdecision}$\blacktriangle$ & Kev-0.8B & joint & 37.9 & 33.8 & 41.0 & 21.3 & 20.0 & 46.2\rlap{$^{\ddagger}$} & \underline{20.3} & 30.0 & 12.0 & 16.4\rlap{$^{\ddagger}$} & 27.9 & 24.9 \\
\rowcolor{rowdecision}$\triangle$ & Kev-0.8B & isolated & 41.8 & 38.4 & 47.9 & 24.7 & 23.9 & \underline{68.1}\rlap{$^{\dagger}$} & 19.2 & \underline{31.5} & \underline{16.5} & \underline{29.9}\rlap{$^{\dagger}$} & \underline{34.2} & \underline{38.2} \\
\rowcolor{rowreranker}$\blacksquare$ & Prism-0.8B &  & \textbf{72.3}\rlap{$^{\ast}$} & \textbf{67.7}\rlap{$^{\ast}$} & \textbf{66.3}\rlap{$^{\ast}$} & \textbf{47.0}\rlap{$^{\ast}$} & \textbf{43.7}\rlap{$^{\ast}$} & \textbf{84.2}\rlap{$^{\ast}$} & \textbf{36.1}\rlap{$^{\ast}$} & \textbf{76.3}\rlap{$^{\ast}$} & \textbf{35.6}\rlap{$^{\ast}$} & \textbf{41.7}\rlap{$^{\ast}$} & \textbf{57.1} & \textbf{53.8} \\
\midrule
\multicolumn{15}{l}{\textit{Qwen3.5-4B}} \\
\rowcolor{rowbase}$\circ$ & Qwen3.5-4B &  & \underline{67.9} & \underline{64.0} & \underline{65.9} & \underline{42.3} & \underline{40.4} & \underline{78.0} & \underline{33.3} & \underline{74.1} & \underline{25.0} & 33.0 & \underline{52.4} & 45.3 \\
\rowcolor{rowdecision}$\blacktriangle$ & Kev-4B & joint & 62.5 & 63.2 & 60.8 & 37.0 & 36.3 & 75.3 & 29.3\rlap{$^{\ddagger}$} & 58.4\rlap{$^{\ddagger}$} & 19.9 & \underline{39.8} & 48.2 & 45.0 \\
\rowcolor{rowdecision}$\triangle$ & Kev-4B & isolated & 62.7 & 62.9 & 60.9 & 37.9 & 34.5\rlap{$^{\dagger}$} & \underline{78.0} & 32.0 & 67.7\rlap{$^{\dagger}$} & 23.4 & 39.0\rlap{$^{\dagger}$} & 49.9 & \underline{46.8} \\
\rowcolor{rowreranker}$\blacksquare$ & Prism-4B &  & \textbf{74.4}\rlap{$^{\ast}$} & \textbf{68.9}\rlap{$^{\ast}$} & \textbf{70.3}\rlap{$^{\ast}$} & \textbf{50.1}\rlap{$^{\ast}$} & \textbf{46.2}\rlap{$^{\ast}$} & \textbf{84.6}\rlap{$^{\ast}$} & \textbf{37.4}\rlap{$^{\ast}$} & \textbf{76.9}\rlap{$^{\ast}$} & \textbf{34.3}\rlap{$^{\ast}$} & \textbf{43.7}\rlap{$^{\ast}$} & \textbf{58.7} & \textbf{54.2} \\
\midrule
\multicolumn{15}{l}{\textit{Qwen3.5-9B}} \\
\rowcolor{rowbase}$\circ$ & Qwen3.5-9B &  & 67.3 & 65.9 & 66.1 & 43.7 & 39.9 & 77.8 & 33.3 & 76.0 & 26.3 & 34.5 & 53.1 & 46.2 \\
\rowcolor{rowdecision}$\blacktriangle$ & Kev-9B & joint & 62.2 & 62.3 & 60.5 & 37.7 & 36.5 & 76.8 & 31.4 & 56.0\rlap{$^{\ddagger}$} & 23.7 & 36.1 & 48.3 & 45.5 \\
\rowcolor{rowdecision}$\triangle$ & Kev-9B & isolated & 63.3 & 64.0 & 60.1 & 38.7 & 34.8 & 75.9 & 33.3 & 71.7 & 21.9 & 35.6 & 49.9 & 44.5 \\
\rowcolor{rowdecision}$\blacktriangle$ & Clef-Flash & joint & 67.5 & 67.4 & 63.9 & 42.9 & 41.2 & 80.9 & 34.1\rlap{$^{\ddagger}$} & 62.4\rlap{$^{\ddagger}$} & 20.0 & 37.5\rlap{$^{\ddagger}$} & 51.8 & 46.1 \\
\rowcolor{rowdecision}$\triangle$ & Clef-Flash & isolated & 67.8 & 69.6\rlap{$^{\dagger}$} & 64.5 & 43.8 & 43.1 & 80.5 & 36.1\rlap{$^{\dagger}$} & 68.1\rlap{$^{\dagger}$} & 23.4 & 40.8\rlap{$^{\dagger}$} & 53.8 & 48.3 \\
\rowcolor{rowreranker}$\blacksquare$ & Prism-9B &  & \textbf{74.0}\rlap{$^{\ast}$} & \textbf{71.1} & \textbf{71.0}\rlap{$^{\ast}$} & \underline{51.6}\rlap{$^{\ast}$} & \underline{44.2} & \underline{83.0} & \textbf{37.2} & \textbf{79.8}\rlap{$^{\ast}$} & \underline{34.7}\rlap{$^{\ast}$} & \underline{43.2}\rlap{$^{\ast}$} & \underline{59.0} & \underline{53.7} \\
\rowcolor{rowreranker}$\blacksquare$ & \textit{Ours-9B} &  & \underline{73.8}\rlap{$^{\ast}$} & \underline{70.0} & \underline{67.3} & \textbf{52.1}\rlap{$^{\ast}$} & \textbf{47.0} & \textbf{84.2} & \underline{37.1} & \underline{77.9}\rlap{$^{\ast}$} & \textbf{36.5}\rlap{$^{\ast}$} & \textbf{47.7}\rlap{$^{\ast}$} & \textbf{59.4} & \textbf{56.1} \\
\midrule
\multicolumn{15}{l}{\textit{Qwen3.8-27B}} \\
\rowcolor{rowbase}$\circ$ & Qwen3.8-27B &  & 71.9 & 68.7 & 66.9 & 45.8 & 44.7 & 82.6 & 37.2 & 74.5 & 30.1 & 41.7 & 56.4 & 51.5 \\
\rowcolor{rowdecision}$\blacktriangle$ & Kev-27B & joint & 72.8 & \textbf{72.9}\rlap{$^{\dagger}$} & \textbf{70.7}\rlap{$^{\dagger}$} & \underline{50.9}\rlap{$^{\dagger}$} & \textbf{49.2}\rlap{$^{\dagger}$} & \textbf{83.9} & \textbf{38.4}\rlap{$^{\dagger}$} & \textbf{81.1}\rlap{$^{\dagger}$} & \textbf{36.1}\rlap{$^{\dagger}$} & 44.9\rlap{$^{\dagger}$} & \textbf{60.1} & \textbf{55.0} \\
\rowcolor{rowdecision}$\triangle$ & Kev-27B & isolated & \underline{73.6} & 70.5 & \underline{70.1} & 50.5 & 46.7 & \underline{83.2} & 37.9 & \underline{80.6} & 32.1 & \underline{45.2} & 59.0 & 53.5 \\
\rowcolor{rowdecision}$\blacktriangle$ & Clef & joint & 70.5 & \underline{72.4} & 66.2 & 48.2 & 45.9 & 80.7 & 35.6 & 58.7 & 29.2 & 40.9 & 54.8 & 50.3 \\
\rowcolor{rowdecision}$\triangle$ & Clef & isolated & 69.4 & 68.5 & 66.5 & 43.7 & 43.3 & 80.3 & 36.2 & 62.7 & 28.1 & 40.5 & 53.9 & 49.6 \\
\rowcolor{rowreranker}$\blacksquare$ & \textit{Ours-27B} &  & \textbf{74.6} & 69.6 & 69.0 & \textbf{52.3} & \underline{48.3} & 83.0 & \underline{38.0} & 79.2 & \underline{34.6} & \textbf{47.1}\rlap{$^{\ast}$} & \underline{59.6} & \underline{54.9} \\
\bottomrule
\end{tabular}
\end{adjustbox}
\end{table*}

A decision model takes an input, the \emph{state}, together with one or more questions about it, and returns a
probability for each possible answer. We use binary questions only, so every question yields one number, the
probability of ``yes''. Reranking follows directly: for a query $q$ with candidates $d_1, \dots, d_k$ from a
first-stage retriever, we ask about each passage $d_i$ whether it answers $q$ and order the passages by the returned
probabilities $p_i$.

For the state, we consider two modes, the same as S1Rank~\cite{S1Rank}. In \emph{isolated} mode, the state holds the
query and a single passage, so that $p_i = P(\textit{yes} \mid q, d_i)$, as in a cross-encoder. In \emph{joint} mode,
the state holds the query and all $k$ candidates in first-stage order, and the $k$ questions are asked about this one
state, so that $p_i = P(\textit{yes} \mid q, d_1, \dots, d_k)$ can depend on the other candidates. Joint mode sees the
same input as a listwise reranker such as RankGPT, but it returns a score per passage instead of an order, like the
Set-Encoder~\cite{Schlatt2025SetEncoder}. In both modes the model scores each passage on its own scale, so both are
pointwise rerankers.

\paragraph{Terms.}
A \emph{backbone} is one pretrained Qwen model of a given size. For every backbone we compare three variants, all of
which return a probability of relevance per query and passage:
\begin{itemize}
  \item The \emph{untuned model} is the released chat model of the backbone, not fine-tuned by anyone for ranking or
  for decisions. We ask it the decision models' question in its chat format and take $p_i$ as the probability of
  ``yes'' as its first answer token.
  \item A \emph{decision fine-tune} (Kev, Clef) is the backbone fine-tuned to act as a decision model.
  \item A \emph{reranker fine-tune} (Qwen3-Reranker, Prism, ours) is the backbone fine-tuned on relevance labels. It
  keeps its own prompt with $q$ and $d_i$ and gives $p_i$ as the probability of ``yes'' against ``no''.
\end{itemize}
We can therefore rank by $p_i$ (RQ1) and read it on an absolute scale (RQ2) for every variant. Decision
models are all models of this kind, including \jev{} (closed) and \laya{}, which we report for context only.

\section{Experimental Setup}
\label{sec:setup}

\begin{table*}[!t]
\caption{Context: other rerankers (nDCG@10, $\times 100$) on the BM25 top-100 of TREC DL 2019--2023 and
five BEIR collections; System One models in joint mode (\laya{} isolated), asked whether the passage answers
the query (Sect.~\ref{sec:setup}). $^\dagger$: published number on
the same candidates~\cite{Abdallah2025HowGood,SetEncoderCard}, not recomputed. Avg: mean over all ten
collections (only where all are available). \$/1k: mean API cost in US dollars per 1,000 queries over the
collections run. Best reranker per column in \texttt{bold}, second best \underline{underlined}.}
\label{tab:context}
\centering
\scriptsize
\setlength{\tabcolsep}{3pt}
\begin{adjustbox}{max width=\textwidth}
\begin{tabular}{llrrrrrrrrrrrr}
\toprule
 & & \multicolumn{5}{c}{TREC DL} & \multicolumn{5}{c}{BEIR} & & \\
\cmidrule(lr){3-7}\cmidrule(lr){8-12}
System & Mode & DL19 & DL20 & DL21 & DL22 & DL23 & COVID & NFC & SciF & Touch\'e & DBP & Avg & \$/1k \\
\midrule
BM25 &  & 50.6 & 48.0 & 44.6 & 26.9 & 26.2 & 59.5 & 32.2 & 67.9 & 44.2 & 31.8 & 43.2 & -- \\
\midrule
\multicolumn{14}{l}{\textit{Listwise LLM rerankers}} \\
RankGPT-4 &  & \textbf{75.6}\rlap{$^\dagger$} & 70.6\rlap{$^\dagger$} & -- & -- & -- & \textbf{85.5}\rlap{$^\dagger$} & \underline{38.5}\rlap{$^\dagger$} & 75.0\rlap{$^\dagger$} & 38.6\rlap{$^\dagger$} & \textbf{47.1}\rlap{$^\dagger$} & -- & -- \\
RankZephyr-7B &  & \underline{74.3} & 70.0 & 67.9 & 50.5 & 44.1 & 80.7\rlap{$^\dagger$} & 36.6\rlap{$^\dagger$} & 75.1\rlap{$^\dagger$} & 31.1\rlap{$^\dagger$} & 43.2\rlap{$^\dagger$} & 57.3 & -- \\
\midrule
\multicolumn{14}{l}{\textit{Cross-encoders and LLM rerankers}} \\
bge-reranker-v2-m3 &  & 72.2\rlap{$^\dagger$} & 67.0\rlap{$^\dagger$} & 66.1 & \underline{51.5} & 45.1 & 74.8\rlap{$^\dagger$} & 33.8\rlap{$^\dagger$} & 73.5\rlap{$^\dagger$} & \underline{39.9}\rlap{$^\dagger$} & 41.9\rlap{$^\dagger$} & 56.6 & -- \\
monoT5-3B &  & 71.8 & 68.9 & 66.7 & 49.5 & 45.0 & 80.7\rlap{$^\dagger$} & \textbf{39.0}\rlap{$^\dagger$} & 76.6\rlap{$^\dagger$} & 32.4\rlap{$^\dagger$} & 44.5\rlap{$^\dagger$} & 57.5 & -- \\
Qwen3-Reranker-8B &  & 73.2 & 66.8 & 67.0 & 51.2 & 45.2 & \underline{84.0} & 37.3 & 79.9 & \textbf{43.1} & 46.2 & \underline{59.4} & -- \\
Set-Encoder-large &  & 75.0\rlap{$^\dagger$} & 72.2\rlap{$^\dagger$} & \underline{69.9} & \textbf{53.4} & \underline{47.7} & 81.2 & 36.8 & 73.8 & 33.2 & \underline{47.0} & 59.0 & -- \\
\midrule
\multicolumn{14}{l}{\textit{System One models}} \\
\jev{} & joint & 73.0 & 70.9 & \textbf{70.7} & 50.3 & 46.1 & 83.3 & 37.8 & \underline{80.3} & 33.5 & 45.5 & 59.1 & 1.11 \\
Clef-Flash & joint & 67.5 & 67.4 & 63.9 & 42.9 & 41.2 & 80.9 & 34.1 & 62.4 & 20.0 & 37.5 & 51.8 & -- \\
Clef & joint & 70.5 & \underline{72.4} & 66.2 & 48.2 & 45.9 & 80.7 & 35.6 & 58.7 & 29.2 & 40.9 & 54.8 & -- \\
Kev-4B & joint & 62.5 & 63.2 & 60.8 & 37.0 & 36.3 & 75.3 & 29.3 & 58.4 & 19.9 & 39.8 & 48.2 & -- \\
Kev-27B & joint & 72.8 & \textbf{72.9} & \textbf{70.7} & 50.9 & \textbf{49.2} & 83.9 & 38.4 & \textbf{81.1} & 36.1 & 44.9 & \textbf{60.1} & -- \\
\laya{} & isolated & 59.5 & 54.6 & 52.8 & 35.3 & 32.7 & 73.8 & 28.7 & 44.8 & 21.8 & 31.3 & 43.5 & -- \\
\bottomrule
\end{tabular}
\end{adjustbox}
\end{table*}

\paragraph{Backbones and variants.}
We study six Qwen backbones: Qwen3-0.6B and 8B, Qwen3.5-0.8B, 4B and 9B, and Qwen3.8-27B, each in the three variants
of Sect.~\ref{sec:decision-models} (Figure~\ref{fig:teaser}). The untuned model is asked with thinking off. The
decision fine-tunes are Kev~\cite{Kev,KevCards} on every backbone, Clef-Flash on the 9B and Clef on the 27B
backbone~\cite{Clef}, each run in isolated and in joint mode. Kev up to 9B is a LoRA fine-tune of the pretrained
checkpoint; Kev-27B is a full fine-tune of the released chat model. The decision fine-tunes are trained to give
\jev{}'s kind of output; they are not reproductions of RLCD, which is unpublished~\cite{KevCards}. The reranker
fine-tunes are Qwen3-Reranker~\cite{Zhang2025Qwen3Embedding} (0.6B, 8B), built on the same pretrained checkpoints as
Kev, and Prism~\cite{Zhang2026Prism} (0.8B, 4B, 9B).

\paragraph{Our reranker fine-tunes.}
No reranker fine-tune of Qwen3.8-27B has been released, so we train one ourselves (\emph{Ours-27B}), from exactly the
checkpoint that Kev-27B and Clef start from. As a control, we train a second one with the same recipe on Qwen3.5-9B
(\emph{Ours-9B}), where it can be compared with Prism-9B. The recipe is public and deliberately plain: LoRA ($r=32$)
in Qwen3-Reranker's prompt format, one epoch over 100k MS~MARCO training queries~\cite{Nguyen2016MSMARCO}, each with
one relevant passage and 15 hard negatives (eight from a dense retriever, seven from BM25), and a ranking loss
(InfoNCE over the group) plus a yes/no classification loss (binary cross-entropy of every passage's score). The second
term gives the model a meaningful $P(\textit{yes})$. We removed training queries that match a test or development
topic and chose the checkpoint on 200 MS~MARCO dev queries by a rule fixed in advance. No decision data and no
hyperparameter search are involved. Qwen3-Reranker, Prism and our fine-tunes saw MS~MARCO in training; Kev's training
data contain neither MS~MARCO nor BEIR~\cite{KevCards}.

\paragraph{The question.}
Every decision and untuned model is asked one question per passage: ``Does document \texttt{\{doc\}} answer the
\texttt{query}?'' It matches the relevance labels we evaluate against (on TREC DL, a passage counts as relevant if it
answers the query, from grade~2). We fixed this question before running the experiments, but did not choose it on
development data, and we report every result with it. The only other wording we ran is the question of
S1Rank~\cite{S1Rank}: ``Is document \texttt{\{doc\}} relevant to the \texttt{query}? Relevant means it answers the
query or contains information that directly addresses it.'' Section~\ref{sec:calibration} compares the two.
Reranker fine-tunes ask no question of this kind; they keep their own prompt.

\paragraph{Collections and candidates.}
We use TREC DL
2019--2023~\cite{Craswell2020DL19,Craswell2021DL20,Craswell2021DL21,Craswell2022DL22,Craswell2023DL23} on the
MS~MARCO v1 and v2 passages~\cite{Nguyen2016MSMARCO} and five BEIR collections~\cite{Thakur2021BEIR}: TREC-COVID,
NFCorpus, SciFact, Touch\'e and DBPedia. None of the fine-tunes was trained on TREC-COVID, Touch\'e or DBPedia, which
form the \emph{zero-shot subset} (ZS). Every system reranks the BM25 top-100 of Pyserini~\cite{Lin2021Pyserini}; these
are the candidates of S1Rank~\cite{S1Rank} and of earlier
comparisons~\cite{Sun2023RankGPT,Pradeep2023RankZephyr,Abdallah2025HowGood}, whose published numbers we use where we
did not rerun a system.

\paragraph{Other systems.}
For context, we report BM25, \jev{} 1.13.0 (closed, in both modes), \laya{}~\cite{Laya} (isolated only, as its
context is too short for a joint state), monoT5-3B~\cite{Nogueira2020monoT5}, bge-reranker-v2-m3~\cite{Chen2024BGEM3},
the Set-Encoder~\cite{Schlatt2025SetEncoder}, RankZephyr~\cite{Pradeep2023RankZephyr} (run by us on TREC DL) and
RankGPT-4~\cite{Sun2023RankGPT} (published numbers only).

\paragraph{Ranking.}
We measure nDCG@10 and test differences with paired $t$-tests over queries, Holm-corrected across the ten
collections ($p < 0.05$). Per backbone, we test three differences: the reranker fine-tune against the strongest decision
fine-tune (the decision fine-tune and mode with the best average), that decision fine-tune against the untuned model,
and joint against isolated mode.

\paragraph{Calibration and its uses.}
We compute the expected calibration error (ECE, 15 equal-width bins)~\cite{Guo2017Calibration} and the Brier score
(the mean squared error of the probability against the binary label, which also rewards separating relevant from
non-relevant passages) on the judged query--passage pairs of TREC DL 2019--2023, the only collections that also judge
many non-relevant passages; a passage counts as relevant from grade~2. We test two uses of the probabilities. The \emph{fixed cutoff} keeps every candidate with $p \ge 0.5$; we report the
$F_1$ of this set (unjudged passages count as non-relevant) and its size. Its \emph{regret} is what a better threshold
would gain: the $F_1$ at the best threshold of each collection, chosen afterwards, minus the $F_1$ at 0.5. The
\emph{count test} adds up a query's probabilities over its judged passages. If the probabilities are calibrated, this
sum estimates how many of the passages are relevant. We report the mean absolute error of this estimate, in
passages. As a reference, always guessing the median number of relevant passages of the collection is off by 8.6
passages. Reranker fine-tunes are also calibrated after the fact, with
temperature and Platt scaling of their logit fitted on 1,000 MS~MARCO dev queries, never on a test collection.

\paragraph{Sanity checks.}
Our BM25, monoT5-3B and RankZephyr runs reproduce published numbers~\cite{Pyserini2CR,Abdallah2025HowGood} within 0.2
points nDCG@10. \jev{} in joint mode, asked S1Rank's question, reproduces S1Rank's numbers~\cite{S1Rank} within 0.5
points on four of five collections (1.5 below on TREC-COVID).

\section{RQ1: Decision vs.\ Reranker Fine-Tuning}
\label{sec:reranking}

\begin{table*}[!t]
\caption{Calibration and its uses on the judged pairs of TREC DL 2019--2023 (means). Per backbone:
untuned model, decision fine-tunes in their better-calibrated mode (j/i: joint/isolated), reranker fine-tunes.
ECE (lower is better): N native; T, P after temperature/Platt scaling fitted on MS~MARCO dev (rerankers
only). Brier: mean squared error of the probability. $F_1$@.5 ($\times 100$): set kept at $p \ge 0.5$, native
(N) and Platt-scaled (P); Regret: $F_1$ at each collection's best cutoff minus $F_1$@.5; $|S|$: passages kept
at 0.5 (of 100). Count: mean absolute error of the summed probabilities against the number of relevant
passages (always guessing the median: 8.6). Per backbone and column: best in bold, second best underlined.}
\label{tab:calibration}
\centering
\scriptsize
\setlength{\tabcolsep}{2.2pt}
\begin{adjustbox}{max width=\textwidth}
\begin{tabular}[t]{clrrrrrrrrr}
\toprule
 & & \multicolumn{3}{c}{ECE} & & \multicolumn{2}{c}{$F_1$@.5} & & & \\
\cmidrule(lr){3-5}\cmidrule(lr){7-8}
 & System & N & T & P & Brier & N & P & Regret & $|S|$ & Count \\
\midrule
\multicolumn{11}{l}{\textit{Qwen3-0.6B}} \\
\rowcolor{rowbase}$\circ$ & Qwen3-0.6B & 0.750 & -- & -- & 0.750 & 19.9 & -- & \textbf{0.5} & 100.0 & 36.5 \\
\rowcolor{rowdecision}$\triangle$ & Kev-0.6B (i) & \textbf{0.503} & -- & -- & \textbf{0.463} & \underline{21.4} & -- & \underline{2.2} & 79.8 & \textbf{24.7} \\
\rowcolor{rowreranker}$\blacksquare$ & Qwen3-Reranker-0.6B & \underline{0.511} & \textbf{0.326} & \textbf{0.226} & \underline{0.481} & \textbf{31.8} & \textbf{6.3} & 14.9 & 58.4 & \underline{24.8} \\
\midrule
\multicolumn{11}{l}{\textit{Qwen3-8B}} \\
\rowcolor{rowbase}$\circ$ & Qwen3-8B & \underline{0.231} & -- & -- & \underline{0.233} & \underline{42.5} & -- & \textbf{2.1} & 22.9 & \underline{8.7} \\
\rowcolor{rowdecision}$\blacktriangle$ & Kev-8B (j) & \textbf{0.113} & -- & -- & \textbf{0.177} & 34.0 & -- & \underline{3.1} & 13.7 & \textbf{7.6} \\
\rowcolor{rowreranker}$\blacksquare$ & Qwen3-Reranker-8B & 0.257 & \textbf{0.237} & \textbf{0.227} & 0.238 & \textbf{46.3} & \textbf{7.4} & 9.0 & 32.2 & 12.9 \\
\midrule
\multicolumn{11}{l}{\textit{Qwen3.5-0.8B}} \\
\rowcolor{rowbase}$\circ$ & Qwen3.5-0.8B & 0.617 & -- & -- & 0.566 & 21.6 & -- & 11.2 & 88.8 & 29.9 \\
\rowcolor{rowdecision}$\triangle$ & Kev-0.8B (i) & \underline{0.336} & -- & -- & \underline{0.297} & \underline{25.1} & -- & \underline{2.9} & 51.9 & \underline{16.9} \\
\rowcolor{rowreranker}$\blacksquare$ & Prism-0.8B & \textbf{0.243} & \textbf{0.229} & \textbf{0.237} & \textbf{0.193} & \textbf{46.9} & \textbf{5.7} & \textbf{2.8} & 26.5 & \textbf{12.7} \\
\midrule
\multicolumn{11}{l}{\textit{Other systems}} \\
\rowcolor{rowdecision}$\blacktriangle$ & \jev{} (j) & \textbf{0.075} & -- & -- & \textbf{0.127} & \textbf{54.7} & -- & \underline{1.7} & 16.0 & \textbf{6.3} \\
\rowcolor{rowdecision}$\triangle$ & \laya{} & \underline{0.127} & -- & -- & \underline{0.183} & 37.5 & -- & \textbf{1.0} & 23.3 & \underline{9.2} \\
\rowcolor{rowreranker}$\blacksquare$ & monoT5-3B & 0.312 & 0.257 & \textbf{0.232} & 0.295 & 42.0 & \textbf{5.8} & 9.9 & 35.6 & 15.2 \\
\rowcolor{rowreranker}$\blacksquare$ & bge-reranker-v2-m3 & 0.231 & \textbf{0.226} & \underline{0.236} & 0.228 & \underline{48.2} & \underline{5.7} & 3.8 & 29.3 & 12.1 \\
\rowcolor{rowreranker}$\blacksquare$ & Set-Encoder-large & 0.495 & \underline{0.250} & \underline{0.236} & 0.426 & 31.6 & \underline{5.7} & 21.1 & 52.2 & 24.3 \\
\bottomrule
\end{tabular}
\quad
\begin{tabular}[t]{clrrrrrrrrr}
\toprule
 & & \multicolumn{3}{c}{ECE} & & \multicolumn{2}{c}{$F_1$@.5} & & & \\
\cmidrule(lr){3-5}\cmidrule(lr){7-8}
 & System & N & T & P & Brier & N & P & Regret & $|S|$ & Count \\
\midrule
\multicolumn{11}{l}{\textit{Qwen3.5-4B}} \\
\rowcolor{rowbase}$\circ$ & Qwen3.5-4B & \underline{0.181} & -- & -- & 0.195 & \underline{44.3} & -- & 3.1 & 27.1 & \underline{9.8} \\
\rowcolor{rowdecision}$\triangle$ & Kev-4B (i) & \textbf{0.132} & -- & -- & \underline{0.175} & 42.6 & -- & \textbf{1.4} & 21.9 & \textbf{8.4} \\
\rowcolor{rowreranker}$\blacksquare$ & Prism-4B & 0.218 & \textbf{0.202} & \textbf{0.238} & \textbf{0.172} & \textbf{51.0} & \textbf{5.7} & \underline{2.8} & 23.1 & 11.5 \\
\midrule
\multicolumn{11}{l}{\textit{Qwen3.5-9B}} \\
\rowcolor{rowbase}$\circ$ & Qwen3.5-9B & 0.165 & -- & -- & 0.181 & 46.5 & -- & 2.9 & 24.4 & 8.7 \\
\rowcolor{rowdecision}$\blacktriangle$ & Kev-9B (j) & 0.160 & -- & -- & 0.177 & 40.2 & -- & \underline{2.3} & 23.8 & 9.5 \\
\rowcolor{rowdecision}$\blacktriangle$ & Clef-Flash (j) & \underline{0.112} & -- & -- & \underline{0.158} & \underline{47.1} & -- & \textbf{1.7} & 18.9 & \underline{8.1} \\
\rowcolor{rowreranker}$\blacksquare$ & Prism-9B & 0.242 & \underline{0.233} & \underline{0.238} & 0.183 & \textbf{50.7} & \underline{5.7} & 4.0 & 26.0 & 12.3 \\
\rowcolor{rowreranker}$\blacksquare$ & \textit{Ours-9B} & \textbf{0.111} & \textbf{0.108} & \textbf{0.231} & \textbf{0.157} & 46.6 & \textbf{6.9} & 6.4 & 8.7 & \textbf{7.3} \\
\midrule
\multicolumn{11}{l}{\textit{Qwen3.8-27B}} \\
\rowcolor{rowbase}$\circ$ & Qwen3.8-27B & 0.117 & -- & -- & 0.144 & \underline{54.1} & -- & \underline{1.9} & 16.0 & \textbf{5.8} \\
\rowcolor{rowdecision}$\blacktriangle$ & Kev-27B (j) & \textbf{0.076} & -- & -- & \textbf{0.130} & \textbf{57.2} & -- & \textbf{0.9} & 16.1 & \underline{6.0} \\
\rowcolor{rowdecision}$\blacktriangle$ & Clef (j) & \underline{0.091} & -- & -- & \underline{0.139} & 53.4 & -- & 2.2 & 17.7 & 6.7 \\
\rowcolor{rowreranker}$\blacksquare$ & \textit{Ours-27B} & 0.110 & \textbf{0.114} & \textbf{0.230} & 0.153 & 48.3 & \textbf{8.1} & 5.9 & 7.9 & 7.3 \\
\bottomrule
\end{tabular}
\end{adjustbox}
\end{table*}

\paragraph{Reranker fine-tuning beats decision fine-tuning.}
On every backbone from 0.6B to 9B, the reranker fine-tune ranks better than the decision fine-tune in either mode
(Table~\ref{tab:triplets}). Averaged over the ten collections, its margin over the better decision mode ranges from
5.2 points nDCG@10 at Qwen3.5-9B (Prism-9B 59.0 vs.\ Clef-Flash 53.8) to 28.8 points at Qwen3-0.6B (56.1 vs.\ 27.3).
The difference is significant on 48 of the 60 pairs of reranker fine-tune and collection, and the decision fine-tune
is never significantly better (${}^\ast$). The margin does not come from training on MS~MARCO: on the zero-shot
subset, which no fine-tune was trained on, it ranges from 5.4 to 23.9 points. Nor does it come from the undisclosed
recipes of the published rerankers: Ours-9B, trained on MS~MARCO alone with standard losses, matches Prism-9B (59.4
vs.\ 59.0 on average, 56.1 vs.\ 53.7 on the zero-shot subset). At 27B, the two kinds of fine-tuning tie. Our reranker
fine-tune of the checkpoint that Kev-27B starts from reaches 59.6 against Kev-27B's 60.1, and 54.9 against 55.0 on
the zero-shot subset; they differ significantly on one collection only, DBPedia, where ours is better. On no backbone
does decision fine-tuning make the better reranker.

\paragraph{Up to 9B, decision fine-tuning adds nothing over the untuned model.}
Up to 9B, the better mode of a decision fine-tune stays within $-2.5$ and $+0.7$ points of its untuned model, asked
the same question, with significant gains on at most three of ten collections (${}^\dagger$). Reranker
fine-tuning of the same backbones adds 5.9 (Prism-9B) to 29.5 points (Qwen3-Reranker-0.6B). A user who takes the
released Qwen3.5-4B or Qwen3.5-9B and asks it whether a passage answers the query gets the same ranking as from the
decision fine-tunes built on it, at no training cost. At 27B, both kinds of fine-tuning add about the same to the untuned
model (56.4): Kev-27B 3.7 points, significantly on eight collections, and our reranker fine-tune 3.2. Clef, the other
27B decision fine-tune, adds nothing (54.8 joint, 53.9 isolated).

\paragraph{Context.}
Table~\ref{tab:context} places the triplets among other rerankers on the same candidates. \jev{} (59.1) ranks as well
as the reranker fine-tunes of 8B and 9B (59.0 to 59.4) and the Set-Encoder (59.0). The open Kev-27B (60.1) matches
the closed model it imitates, and so does our 27B reranker fine-tune (59.6). A fine-tune of a strong open backbone
reaches \jev{}, with or without decision data. \laya{} (43.5) does not improve on
its BM25 candidates (43.2).

\paragraph{Joint scoring helps only the largest models.}
Scoring all candidates in one state gains about one point for Kev-27B (60.1 vs.\ 59.0) and Clef (54.8 vs.\ 53.9), but
costs Kev-0.6B 12.2 points and Kev-8B 13.0 (${}^\ddagger$; significant on all ten and on five collections). A likely
reason is the length of the state. A joint state of 100 passages has about 13k tokens on TREC DL and 40k or more on
BEIR, whereas the model cards of the small Kevs report training states of at most 7,552 tokens (Kev-0.8B and Kev-4B),
and that of Kev-27B states of up to 32,768~\cite{KevCards}.

\section{RQ2: Are Decision Models Better Calibrated?}
\label{sec:calibration}

\begin{table}[!t]
\caption{Ablation study on the question wording: S1Rank's question (\emph{Rel.}: ``Is document \{doc\} relevant to the
query? Relevant means it answers the query or contains information that directly addresses it.'') against
the question used in the rest of the paper (\emph{Ans.}: ``Does document \{doc\} answer the query?''), for
every system that is asked a question. nDCG@10 ($\times 100$): mean over all ten collections (Avg) and over
TREC DL 2019--2023 (DL); ECE and Count as in Table~\ref{tab:calibration} (DL 2019--2023). $\Delta$: Ans.\ minus
Rel. Training: $\circ$~untuned model, $\blacktriangle$/$\triangle$~decision fine-tune (joint/isolated), $\blacksquare$~reranker fine-tune.}
\label{tab:question-ablation}
\centering
\scriptsize
\setlength{\tabcolsep}{3pt}
\begin{adjustbox}{max width=\textwidth}
\begin{tabular}{cllrrrrrrrrrrrr}
\toprule
 & & & \multicolumn{3}{c}{nDCG@10 Avg} & \multicolumn{3}{c}{nDCG@10 DL} & \multicolumn{3}{c}{ECE} & \multicolumn{3}{c}{Count} \\
\cmidrule(lr){4-6}\cmidrule(lr){7-9}\cmidrule(lr){10-12}\cmidrule(lr){13-15}
 & System & Mode & Rel. & Ans. & $\Delta$ & Rel. & Ans. & $\Delta$ & Rel. & Ans. & $\Delta$ & Rel. & Ans. & $\Delta$ \\
\midrule
\multicolumn{15}{l}{\textit{Qwen3-0.6B}} \\
\rowcolor{rowbase}$\circ$ & Qwen3-0.6B &  & 23.8 & 26.6 & $+$2.8 & 25.4 & 30.7 & $+$5.3 & 0.750 & 0.750 & \phantom{$+$}0.000 & 36.5 & 36.5 & \phantom{$+$}0.0 \\
\rowcolor{rowdecision}$\blacktriangle$ & Kev-0.6B & joint & 13.2 & 15.1 & $+$1.9 & 14.9 & 16.7 & $+$1.8 & 0.591 & 0.503 & $-$0.088 & 28.8 & 24.6 & $-$4.2 \\
\rowcolor{rowdecision}$\triangle$ & Kev-0.6B & isolated & 27.0 & 27.3 & $+$0.3 & 28.8 & 29.5 & $+$0.7 & 0.579 & 0.503 & $-$0.076 & 28.2 & 24.7 & $-$3.5 \\
\midrule
\multicolumn{15}{l}{\textit{Qwen3-8B}} \\
\rowcolor{rowbase}$\circ$ & Qwen3-8B &  & 53.3 & 51.1 & $-$2.2 & 56.9 & 54.0 & $-$2.9 & 0.383 & 0.231 & $-$0.152 & 17.5 & 8.7 & $-$8.8 \\
\rowcolor{rowdecision}$\blacktriangle$ & Kev-8B & joint & 38.8 & 38.7 & $-$0.1 & 50.9 & 49.3 & $-$1.6 & 0.243 & 0.113 & $-$0.130 & 12.8 & 7.6 & $-$5.2 \\
\rowcolor{rowdecision}$\triangle$ & Kev-8B & isolated & 52.7 & 51.7 & $-$1.0 & 54.2 & 53.8 & $-$0.4 & 0.283 & 0.154 & $-$0.129 & 14.1 & 8.7 & $-$5.4 \\
\midrule
\multicolumn{15}{l}{\textit{Qwen3.5-0.8B}} \\
\rowcolor{rowbase}$\circ$ & Qwen3.5-0.8B &  & 37.6 & 33.9 & $-$3.7 & 40.3 & 39.1 & $-$1.2 & 0.716 & 0.617 & $-$0.099 & 34.8 & 29.9 & $-$4.9 \\
\rowcolor{rowdecision}$\blacktriangle$ & Kev-0.8B & joint & 25.3 & 27.9 & $+$2.6 & 28.9 & 30.8 & $+$1.9 & 0.395 & 0.398 & $+$0.003 & 19.5 & 19.6 & $+$0.1 \\
\rowcolor{rowdecision}$\triangle$ & Kev-0.8B & isolated & 36.0 & 34.2 & $-$1.8 & 36.1 & 35.3 & $-$0.8 & 0.357 & 0.336 & $-$0.021 & 18.0 & 16.9 & $-$1.1 \\
\midrule
\multicolumn{15}{l}{\textit{Qwen3.5-4B}} \\
\rowcolor{rowbase}$\circ$ & Qwen3.5-4B &  & 52.7 & 52.4 & $-$0.3 & 55.8 & 56.1 & $+$0.3 & 0.296 & 0.181 & $-$0.115 & 14.7 & 9.8 & $-$4.9 \\
\rowcolor{rowdecision}$\blacktriangle$ & Kev-4B & joint & 49.8 & 48.2 & $-$1.6 & 53.4 & 51.9 & $-$1.5 & 0.251 & 0.201 & $-$0.050 & 13.2 & 11.6 & $-$1.6 \\
\rowcolor{rowdecision}$\triangle$ & Kev-4B & isolated & 51.3 & 49.9 & $-$1.4 & 52.7 & 51.8 & $-$0.9 & 0.216 & 0.132 & $-$0.084 & 11.5 & 8.4 & $-$3.1 \\
\midrule
\multicolumn{15}{l}{\textit{Qwen3.5-9B}} \\
\rowcolor{rowbase}$\circ$ & Qwen3.5-9B &  & 53.7 & 53.1 & $-$0.6 & 56.7 & 56.6 & $-$0.1 & 0.283 & 0.165 & $-$0.118 & 14.1 & 8.7 & $-$5.4 \\
\rowcolor{rowdecision}$\blacktriangle$ & Kev-9B & joint & 50.2 & 48.3 & $-$1.9 & 54.0 & 51.9 & $-$2.1 & 0.237 & 0.160 & $-$0.077 & 12.2 & 9.5 & $-$2.7 \\
\rowcolor{rowdecision}$\triangle$ & Kev-9B & isolated & 50.1 & 49.9 & $-$0.2 & 52.8 & 52.2 & $-$0.6 & 0.348 & 0.221 & $-$0.127 & 16.9 & 11.8 & $-$5.1 \\
\rowcolor{rowdecision}$\blacktriangle$ & Clef-Flash & joint & 52.8 & 51.8 & $-$1.0 & 57.0 & 56.6 & $-$0.4 & 0.140 & 0.112 & $-$0.028 & 8.9 & 8.1 & $-$0.8 \\
\rowcolor{rowdecision}$\triangle$ & Clef-Flash & isolated & 54.7 & 53.8 & $-$0.9 & 58.9 & 57.8 & $-$1.1 & 0.173 & 0.144 & $-$0.029 & 9.7 & 8.2 & $-$1.5 \\
\midrule
\multicolumn{15}{l}{\textit{Qwen3.8-27B}} \\
\rowcolor{rowbase}$\circ$ & Qwen3.8-27B &  & 57.4 & 56.4 & $-$1.0 & 59.9 & 59.6 & $-$0.3 & 0.166 & 0.117 & $-$0.049 & 8.9 & 5.8 & $-$3.1 \\
\rowcolor{rowdecision}$\blacktriangle$ & Kev-27B & joint & 61.1 & 60.1 & $-$1.0 & 63.5 & 63.3 & $-$0.2 & 0.156 & 0.076 & $-$0.080 & 8.5 & 6.0 & $-$2.5 \\
\rowcolor{rowdecision}$\triangle$ & Kev-27B & isolated & 59.8 & 59.0 & $-$0.8 & 62.5 & 62.3 & $-$0.2 & 0.262 & 0.180 & $-$0.082 & 12.9 & 9.4 & $-$3.5 \\
\rowcolor{rowdecision}$\blacktriangle$ & Clef & joint & 56.1 & 54.8 & $-$1.3 & 61.1 & 60.6 & $-$0.5 & 0.147 & 0.091 & $-$0.056 & 8.4 & 6.7 & $-$1.7 \\
\rowcolor{rowdecision}$\triangle$ & Clef & isolated & 56.4 & 53.9 & $-$2.5 & 58.5 & 58.3 & $-$0.2 & 0.240 & 0.153 & $-$0.087 & 12.1 & 8.3 & $-$3.8 \\
\midrule
\multicolumn{15}{l}{\textit{Other systems}} \\
\rowcolor{rowdecision}$\blacktriangle$ & \jev{} & joint & 60.5 & 59.1 & $-$1.4 & 62.7 & 62.2 & $-$0.5 & 0.224 & 0.075 & $-$0.149 & 11.3 & 6.3 & $-$5.0 \\
\rowcolor{rowdecision}$\triangle$ & \jev{} & isolated & 59.5 & 57.1 & $-$2.4 & 61.4 & 60.0 & $-$1.4 & 0.260 & 0.125 & $-$0.135 & 12.9 & 7.8 & $-$5.1 \\
\rowcolor{rowdecision}$\triangle$ & \laya{} & isolated & 44.0 & 43.5 & $-$0.5 & 47.3 & 47.0 & $-$0.3 & 0.131 & 0.127 & $-$0.004 & 9.3 & 9.2 & $-$0.1 \\
\bottomrule
\end{tabular}
\end{adjustbox}
\end{table}

Calibration is the main promise of decision models. Table~\ref{tab:calibration} reports it on the judged pairs of TREC
DL 2019--2023, together with the two uses we test. Table~\ref{tab:question-ablation} repeats every run of a decision
or untuned model with S1Rank's question.

\paragraph{Asked our question, decision models are well calibrated.}
The untuned models of 0.6B and 0.8B parameters say ``yes'' to almost everything (ECE 0.750 and 0.617); from 4B on,
they are reasonably calibrated already (0.117 to 0.231). Decision fine-tuning improves on the untuned model on every
backbone, for example from 0.231 to 0.113 on Qwen3-8B. From 4B on, the decision fine-tunes are also better calibrated
than the published reranker fine-tunes (Kev-8B 0.113 vs.\ Qwen3-Reranker-8B 0.257); on the two smallest backbones they
are not. The best calibrated systems in the study are \jev{} (0.075) and Kev-27B (0.076). Decision fine-tuning is not
the only way there: our 9B reranker fine-tune, trained with a plain yes/no loss, is as well calibrated as Clef-Flash
(0.111 vs.\ 0.112). Our 27B reranker fine-tune is less well calibrated than Kev-27B (0.110).

\paragraph{The advantage depends on the question.}
S1Rank's question asks whether the passage is relevant and defines relevance more loosely: the passage ``answers the
query or contains information that directly addresses it''. Both wordings are reasonable, and S1Rank's is the one
under which \jev{} was first shown to be a good reranker. The ranking hardly moves between the two: the mean nDCG@10
of the decision fine-tunes changes by $-2.5$ to $+2.6$ points, and with S1Rank's question, too, the best decision
fine-tune of every backbone up to 9B is 4 points or more behind the reranker fine-tune. The answer to RQ1 does not
depend on the wording. Calibration moves a lot. With S1Rank's question, almost every model is worse calibrated, often
by a factor of two or three (\jev{} joint: ECE 0.224 instead of 0.075). The decision fine-tunes are then at best as
well calibrated as the best reranker fine-tune of their backbone (Kev-8B 0.243 vs.\ Qwen3-Reranker-8B 0.257, Kev-27B
0.156 vs.\ Ours-27B 0.110). One sentence of the prompt decides whether decision models are better calibrated than
reranker fine-tunes.

The untuned models gain as much from our wording as the decision fine-tunes (Qwen3-8B: 0.383 with S1Rank's question,
0.231 with ours). A question that matches the definition of relevance in the labels makes the probabilities better
calibrated, with or without decision fine-tuning. The user of a decision model has to put that definition into words,
and a small change in the words changes the calibration severalfold. A reranker fine-tune has no such knob: its
notion of relevance is fixed by its training labels.

\paragraph{Calibration does not make up for a worse ranking.}
If every passage with $p \ge 0.5$ is kept, a reranker fine-tune keeps the better set on every backbone up to 9B
(e.g.\ $F_1$@.5 of 51.0 for Prism-4B vs.\ 42.6 for Kev-4B), although it is the worse calibrated model on most of
them. The threshold is not the problem: for the decision models, 0.5 is close to the best threshold of each
collection (regret of 0.9 to 3.1 points). A well-placed threshold on a worse ranking still yields the worse set. Where
the rankings are equally good, calibration does pay off: at 27B, the better calibrated Kev-27B keeps the better set
(57.2 vs.\ 48.3), while our fine-tune is underconfident and keeps only 7.9 passages per query.

\paragraph{Counting relevant passages.}
A ranking cannot tell how many passages are relevant; the sum of calibrated probabilities can. Asked our question,
most decision models from 4B on count better than the median guess (8.6 passages off) and better than the published
reranker fine-tunes of their backbone (Kev-8B 7.6 vs.\ Qwen3-Reranker-8B 12.9). Decision fine-tuning is not what makes
this work: the untuned Qwen3.8-27B counts most accurately of all (5.8), and Ours-9B counts better than Clef-Flash (7.3
vs.\ 8.1). With S1Rank's question, no decision model beats the median guess by more than a fraction of a passage
(best: Clef joint, 8.4).

\paragraph{Calibration after the fact.}
Rerankers can also be calibrated after training. Temperature scaling fitted on MS~MARCO dev lowers the ECE of the
worst calibrated rerankers (Qwen3-Reranker-0.6B from 0.511 to 0.326) and changes the others little. Platt scaling
backfires at the fixed cutoff ($F_1$@.5 of 8.1 or less): MS~MARCO dev has about one relevant passage per query, so the
fitted probabilities are far too low for TREC DL, where queries have many more. A calibration holds only for the
notion of relevance and the base rate it was fitted on (Sect.~\ref{sec:discussion}).

\section{Discussion and Limitations}
\label{sec:discussion}

\paragraph{Why decision fine-tuning does not make a reranker.}
A reranker fine-tune is trained on the very task it is tested on: telling relevant from non-relevant passages of a
query, with a ranking loss over hard negatives. The decision fine-tunes we tested learned to answer many kinds of
questions about many kinds of states, and Kev's training data contain no MS~MARCO and no BEIR~\cite{KevCards}. That
general skill transfers to relevance about as well as the untuned model's own instruction following, which is what we
observe up to 9B. The zero-shot subset shows that the gap is not about having seen the test domain: it persists on
collections that neither kind of fine-tune was trained on.

\paragraph{Why 27B is different.}
Kev-27B is the only decision fine-tune that adds to its untuned model. It differs from the smaller Kevs in more than
size: it is a full fine-tune of the released chat model, whereas the smaller Kevs are LoRA fine-tunes of pretrained
checkpoints, and it was trained on longer states (Sect.~\ref{sec:reranking}). We cannot tell which of these matters.
Size alone is not enough: Clef, a decision fine-tune of the same 27B model, adds nothing. Reranker fine-tuning also
adds less at 27B than on the smaller backbones, because the untuned 27B model already ranks well.

\paragraph{What calibration means for relevance.}
A probability of relevance is calibrated only with respect to a definition of relevance and its base rate. Our
results show three ways to move it: the wording of the question, the labels of the
training data (our reranker fine-tunes) and the labels a scaling is fitted on (Platt scaling on MS~MARCO dev,
Sect.~\ref{sec:calibration}). A model that is ``calibrated out of the box'' is calibrated against a notion of
relevance that someone chose, and the user of a decision model has to put that notion into words. For ranking none of
this matters; for a threshold or a count it matters a lot.

\paragraph{Limitations.}
We compare released models, and the fine-tunes of a backbone differ in their training data as well as in their
objective. Our results therefore describe decision fine-tuning as it is available today, not RLCD as such, which is
unpublished and may be done better. The controlled comparison rests on our own reranker fine-tunes at 9B and 27B,
which were trained once, with one recipe and without any hyperparameter search. We measured calibration only on TREC
DL, where non-relevant passages are judged, and at a single relevance grade and a single threshold. Our question was
fixed before the runs but not chosen on development data, and we tried only one other wording. All decision models
are at most a month old, and newer versions may change the picture; we release all raw outputs so that the comparison
can be repeated.

\section{Conclusion}
\label{sec:conclusion}

System One models are not better rerankers. On the same backbone, decision fine-tuning never makes the better
reranker: plain reranker fine-tuning wins by 5 to 29 points nDCG@10 from 0.6B to 9B, and the two tie at 27B. Up to 9B,
decision fine-tuning adds nothing over simply asking the untuned model. The calibration advantage of decision models
depends on the question. A question that matches the relevance labels makes decision and untuned models alike much
better calibrated; with a looser one, decision models lose their advantage over reranker fine-tunes. Up to 9B,
calibration also does not make up for a worse ranking when passages are kept at a fixed threshold. If you need a
reranker, fine-tune a reranker.

\begin{acks}
The project is supported by funds of the Federal Ministry of Agriculture, Food and Regional Identity (BMLEH) based on a decision of the Parliament of the Federal Republic of Germany via the Federal Office for Agriculture and Food (BLE) under the strategy for digitalisation in agriculture.
\end{acks}

\section*{Disclosure}
The authors declare no competing interests.
Furthermore, the authors used generative AI tools for code writing and language editing
and take full responsibility for the final content.

\balance
\newpage
\bibliographystyle{ACM-Reference-Format}
\bibliography{sources}

\end{document}